\documentclass[letterpaper, 10 pt, conference]{ieeeconf}  

\IEEEoverridecommandlockouts                        
\title{\LARGE \bf
Dream What I Can: Sparse Visual Planning \\for Long-Horizon Robot Manipulation
}

\author{Chongkai Gao$^{12}$, Wenlong Huang$^{1}$, Haonan Chen$^{13}$,\\Lin Shao$^{2}$, Yilun Du$^{3}$, Jiajun Wu$^{1}$, Ruohan Zhang$^{4}$, Li Fei-Fei$^{1}$\\ \\ $^{1}$Stanford University \quad $^{2}$National University of Singapore \quad $^{3}$Harvard University \quad $^{4}$Northwestern University} %

\usepackage{bbm}
\usepackage{etoolbox}
\usepackage{multicol}
\usepackage{cite}
\usepackage[bookmarks=true]{hyperref}
\usepackage{graphics} 
\usepackage{times} 
\usepackage{amsmath} 
\usepackage{amssymb}  
\usepackage{svg}
\usepackage{mathrsfs}
\usepackage{amsfonts}
\usepackage{wrapfig}
\usepackage{amsmath}

\usepackage{kantlipsum}
\usepackage{subcaption}
\usepackage{marvosym}

\usepackage{graphicx}
\usepackage{float}
\usepackage{ctable}
\usepackage{cuted}
\usepackage{colortbl}
\usepackage{multirow}
\usepackage{wrapfig}
\usepackage[misc,geometry]{ifsym}
\usepackage{algorithm}
\usepackage{algpseudocode}
\usepackage{xcolor}   
\usepackage{pifont}    

\usepackage{bm}
\usepackage{caption}
\usepackage{setspace}
\let\labelindent\relax
\usepackage{enumitem}
\usepackage{threeparttable}
\newcommand{\TODO}[1][]{\textcolor{red}{\bf [TODO]}}
\setlist[enumerate,1]{itemsep=3pt}

\newcommand{\name}{DreamCan}
\definecolor{formalgreen}{rgb}{0.1, 0.7, 0.1}  
\definecolor{formalred}{rgb}{0.9, 0.2, 0.2}  

\begin{document}

\maketitle
\thispagestyle{empty}
\pagestyle{empty}

\begin{strip}
\centering
\includegraphics[width=\linewidth]{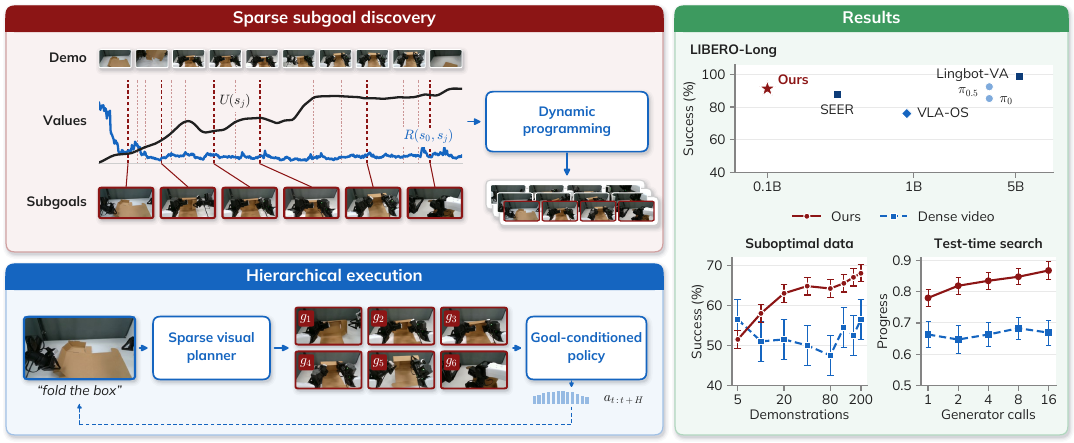}
\captionof{figure}{
\textbf{Overview of \name{}}. Top left: given a video dataset of a manipulation task, our method can perform unsupervised sparse subgoal image sequence discovery to get subgoal sequences that balance subgoal reachability and task utility; Bottom left: trained on these subgoal sequence datasets, the sparse visual planner can generate a subgoal image sequence given an image observation and language instruction, and a goal-conditioned low-level policy can generate robot actions according to the subgoals. Right: with our subgoal sequences, our parameter-efficient low-level policy can achieve comparable performance with large VLAs or WAMs on LIBERO-Long, and can scale with test-time compute and suboptimal data.}
\label{fig:teaser}
\end{strip}

\begin{abstract}
Video models provide rich priors for generalizable robotic manipulation, but plausible visual futures are not always executable. We introduce \textbf{\name}, a hierarchical policy that grounds video models through \emph{sparse visual planning}: a high-level planner adapted from a pre-trained video model and a low-level visuomotor policy. Rather than predicting dense trajectories, the high-level planner generates sparse visual subgoals that balance \emph{reachability} and \emph{task utility}. From offline, action-free robot videos, we estimate these quantities and use dynamic programming to select subgoal sequences without manual annotations. The resulting subgoals are more reliably attained than uniformly spaced alternatives, and the full system outperforms language, dense-video, and uniform-subgoal planners, reaching high planning success rates on Language-Table tasks and 91.2\% success on LIBERO-Long with a 0.1B-parameter low-level policy. Additional suboptimal videos and test-time compute further improve success. The system also performs challenging real-world bimanual tasks, including box folding and packing with unseen object combinations.
More results are on \href{https://dreamcan.github.io/}{https://dreamcan.github.io/}.

\end{abstract}
\section{Introduction}

The promise of video models for robotics lies not merely in predicting how the world may evolve under interaction, but in revealing intermediate outcomes through which a task can be accomplished.
For manipulation, this knowledge extends beyond the task semantics readily expressed in language: visual predictions can depict contact, fine adjustment, and detailed configurations connecting a scene to a desired outcome. Video therefore offers an expressive medium for planning interactions whose geometric and physical details are cumbersome to specify in words.

Yet a plausible visual future is not necessarily one a robot can realize.
We argue that a visual plan should specify intermediate outcomes that support execution, rather than every intermediate frame. A goal image prediction may leave too much for the controller to solve, while a dense trajectory prescribes overly constrained motions that can be hard to realize by the controller. We thus propose \emph{sparse visual planning}: learning to predict a sequence of subgoal images that both (i) makes meaningful progress towards achieving the task and (ii) remains executable by a low-level subgoal-conditioned policy.

We formalize these principles through \emph{reachability}, which estimates how readily one visual state can be attained from another, and \emph{task utility}, which measures progress toward the instruction. Under a fixed subgoal budget, a good subgoal sequence plan should reward terminal utility and penalize difficult transitions. Utility directs the plan toward task completion; reachability determines its intermediate steps. This formulation echoes SayCan’s decomposition into task relevance and robotic affordance~\cite{saycan}, but does not require a predefined vocabulary of language-labeled skills. Instead, intermediate objectives can be discovered from video datasets, allowing them to reflect visual configurations useful for execution rather than language-based task decompositions.

We term our approach \textbf{\name}, a sparse visual planner that generates policy-aware sequences of visual subgoals. Given a dataset of robot manipulation videos, we first learn two complementary value functions from the action-free data: a \textit{reachability value}, estimating whether one visual state can be reliably reached from another, and a \textit{task utility value}, measuring progress toward task completion. We instantiate reachability with goal-conditioned offline RL~\cite{iql,park2023hiql} under sparse rewards, and estimate utility using a task-progress model~\cite{liang2026robometer} or a VLM. We then jointly use these two values to automatically annotate each video with a sequence of visual subgoals that balances local reachability with global task progress. The resulting subgoal sequences are used to fine-tune a pretrained video generation model, WAN-2.2~\cite{wan2025wan}, to predict sparse visual subgoals from the current observation and language instruction, serving as the planning output of our model. These predicted subgoals can be subsequently executed by a goal-conditioned low-level policy.

Through extensive experiments, we show that \name\ consistently outperforms alternative planning representations, including language plans, dense video predictions, and uniformly sampled visual subgoals. On three challenging Language-Table tasks~\cite{lynch2023interactive}, \name\ achieves substantially higher planning success rates than the prior state-of-the-art method~\cite{vlp}. On LIBERO-Long~\cite{liu2023libero}, our hierarchical system achieves a 91.2\% success rate with only a 0.1B-parameter low-level policy. We further show that \name\ benefits from scaling suboptimal action-free video data on RoboMimic-MH Square~\cite{robomimic}, without additional policy training, as well as from increased test-time compute on Language Table~\cite{lynch2023interactive}. The sparse planning representation also naturally supports language-guided steering at test time, enabling users to specify preferred plans. Finally, on a suite of challenging real-world tasks, our system demonstrates effective manipulation under contact-rich interactions, long horizons, and combinatorial generalization using a bimanual robot platform, while maintaining fast closed-loop replanning at 1.03s per planning step on an RTX 5090 GPU.

Our contributions are: (i) We propose \name, a sparse visual planner for robot manipulation that learns from action-free video datasets using automatically discovered subgoals that balance reachability and task utility; (ii) We integrate \name\ with a goal-conditioned policy to form a closed-loop hierarchical manipulation system, and demonstrate strong performance across simulated benchmarks and real-world contact-rich tasks; and (iii) We show that the proposed sparse planning framework scales effectively with both suboptimal video data and test-time compute, while supporting language-guided plan steering at inference time.

\section{Related Work}
\subsection{Keystate Planning for Robot Manipulation Tasks}

Sparse temporal abstractions have long been used to reduce the redundancy of dense robot trajectories. Keystates can be obtained from manual annotations~\cite{akgun2012keyframe,wgobench}, embodiment-specific signals such as actions, proprioception, gripper states, and unsupervised objectives based on temporal structure, reconstruction, or information-theoretic criteria~\cite{gregor2016variational,eysenbach2019diversity,pertsch2020keyin}. However, these approaches either scale poorly or rely on signals unavailable in video-only data. VLMs can instead identify semantically salient task stages~\cite{wang2024vlm}, but semantic salience does not guarantee executability. Several works directly use visual subgoals for planning and control. KeyIn~\cite{pertsch2020keyin} discovers keyframes through prediction and reconstruction, SuSIE~\cite{black2024susie} generates subgoal images through a pretrained image generator, and LBP~\cite{liu2025lbp} predicts latent subgoal sequences using fixed temporal supervision. In contrast, we construct sparse visual subgoals by balancing \emph{task utility} and \emph{reachability}, favoring states that are both reachable and useful.

\subsection{World Models for Robot Manipulation Tasks}

World models support robot control by predicting future dynamics. Latent models such as Dreamer~\cite{hafner2023dreamerv3} and TD-MPC~\cite{hansen2024tdmpc2} enable policy optimization or model-predictive control, while visual models such as UniPi~\cite{du2023unipi}, UniSim~\cite{yang2024unisim}, and other works~\cite{gao2025flip,zhou2026gem} predict future observations for action generation or interaction. More recent world-action models jointly model visual futures and actions~\cite{li2025unifiedvideoaction,ye2026dreamzero}, with Fast-WAM~\cite{yuan2026fastwam} improving inference efficiency. Our work instead focuses on the temporal granularity of explicit visual planning. Rather than predicting dense trajectories, we generate sparse subgoal images supervised by \emph{reachability} and \emph{task utility}. This reduces prediction overhead, enables test-time search, and enables learning from suboptimal datasets. We further decouple the planner and policy, allowing them to communicate solely through subgoal images without sharing gradients or features. This modular design enables either component to be independently replaced or upgraded.

\section{Preliminary and Motivation\label{sec:background}}
\subsection{Preliminaries}
\label{sec:preliminary}

We consider a language-conditioned visuomotor manipulation, where the robot observes an image $o_t \in \mathcal{O}$ and an instruction $\ell \in \mathcal{L}$, and produces a low-level action chunk $a_{t:t+H-1} \in \mathcal{A}$ to complete the task. Rather than predicting actions directly, \name{} adopts a hierarchical architecture. The high-level \emph{sparse visual planner}, denoted by $\mathcal{P}_\theta$, predicts a short sequence of $K$ future visual subgoals:
\begin{equation}
\hat{g}_1, \hat{g}_2, \cdots, \hat{g}_K
=
\mathcal{P}_\theta(o_t,\ell), 
\quad \hat{g}_k \in \mathcal{O},
\end{equation}
which are executed by a goal-conditioned policy $\pi_\phi$:
\begin{equation}
a_{t:t+H-1} \sim
\pi_\phi
\left(
a_{t:t+H-1} \mid o_t,\hat{g}_{1:k},\ell
\right),
\end{equation}
where $\hat{g}_{1:k}$ denotes the actual subgoals used as conditions. The policy repeatedly generates actions toward the current subgoals, while the planner is periodically re-invoked from the latest observation for closed-loop replanning.

Importantly, $\mathcal{P}_\theta$ and $\pi_\phi$ are \emph{decoupled}. The planner is trained solely from video trajectories
$\mathcal{D}_\mathrm{vid}={(\ell^{(i)},o^{(i)}{0})}$, while the low-level policy can be any compatible goal-conditioned policy trained independently. Unlike tightly coupled world-action models, they share neither gradients nor hidden representations; their only interface is the generated subgoal images. This modular design allows either component to be replaced or improved independently.

\subsection{Motivation: What Makes a Good Visual Plan?}

\begin{figure}
    \centering
    \includegraphics[width=\linewidth]{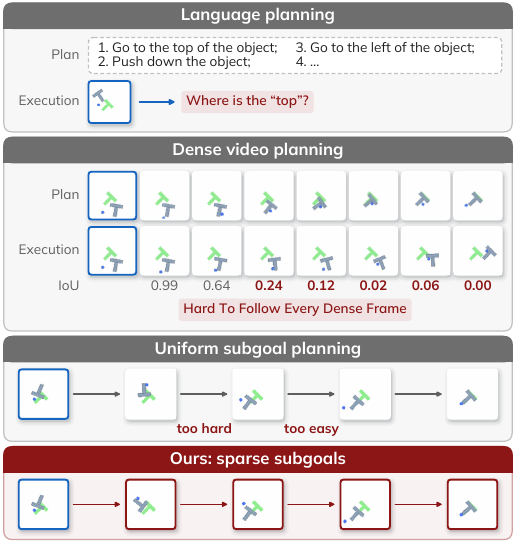}
    \caption{The problems of language planning, dense video planning, and uniform subgoal planning on the Push-T task.}
    \label{fig:motivation}
\end{figure}

We use Push-T~\cite{florence2022implicit} to illustrate why choosing an appropriate planning representation is nontrivial (Figure \ref{fig:motivation}). The robot must push a T-shaped block into a target pose, requiring both geometric precision and appropriate contact changes. 

A natural choice is language. A language planner might say ``\textit{Push the T shape from left to right} " or ``\textit{rotate the T and then move it into place}''. However, such descriptions leaves ambiguous where to contact, which direction to push, and which intermediate configurations to reach.

As an alternative, a dense video planner can predict the entire future trajectory frame by frame, providing richer geometric guidance, but frame-by-frame prediction is prone to accumulated errors, can conflict with the actual execution trajectory after deviations, and is expensive to generate.

A seemingly simple alternative is to uniformly sample a few future frames. However, temporal spacing does not imply good subgoals: selected frames may capture uninformative motion, regress in suboptimal demonstrations, miss important contact changes, or be difficult for the policy to reach.

A useful visual subgoal should balance two properties: \emph{reachability}, whether it can be reliably reached from the preceding states, and \emph{task utility}, whether it makes meaningful progress toward completion. Nearby states may be easy but uninformative, while distant states may be useful but unreachable. This trade-off motivates our formulation for selecting sparse visual plans.

\section{Method\label{sec:method}}

\label{sec:method}

In this section, we first introduce the \emph{reachability} and \emph{task utility} value functions, then use them to annotate sparse visual subgoals for training the sparse visual planner. 

\begin{figure*}[t]
    \centering
    \includegraphics[width=\linewidth]{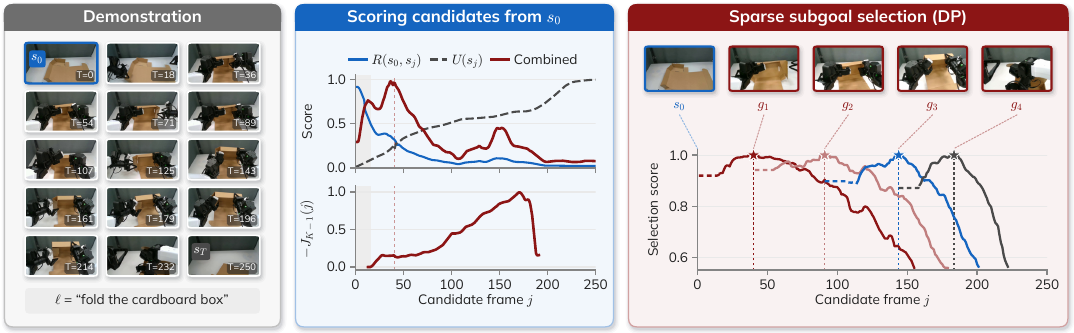}
    \caption{Subgoal annotation for \texttt{Box Folding}.
    Left: demonstration frames.
    Middle: reachability, task utility, their combination, and the dynamic programming cost curves used for selection at the initial frame. Note that reachability and DP curves will change with the starting state.
    Right: full cost curves $J$ during the subgoal selection process.
    We illustrate four subgoals for clarity; the task uses twelve
    in our experiments.}
    \label{fig:method}
\end{figure*}

\subsection{Dataset-Supported Reachability}
\label{sec:reachibility}

\paragraph{Formulation}
We measure transition difficulty using a goal-conditioned value function
$V_R^{\pi}(s_i,s_j)$ with reward $=-1$ before reaching $s_j$ and $0$
at goal termination. For discount $\gamma$ and first hitting
time $\tau_{ij}$, its normalized reachability is:
\begin{equation}
\begin{aligned}
R^{\pi}(s_i,s_j)
&=1+(1-\gamma)V_R^{\pi}(s_i,s_j)\\
&=\mathbb{E}_{\pi}[\gamma^{\tau_{ij}}]\in[0,1],
\end{aligned}
\label{eq:reach-score}
\end{equation}
where $\tau_{ij}=\infty$ if the goal is never reached and
$\gamma^{\infty}=0$. Higher scores indicate faster or more reliable
goal attainment, rather than a success probability at a fixed horizon.

\paragraph{Learning from videos}
Frozen RADIO features~\cite{ranzinger2024radio} encode observations
and goals; a temporal Transformer summarizes observation history and
fuses it with goal and language features to predict $V_R$.
We use action-free goal-conditioned expectile
learning~\cite{park2023hiql,iql}, with expectile $0.75$, discount $0.95$,
Polyak target update rate $0.005$, and terminal boundary
$V_R(s_g,s_g)=0$.
Goals are sampled equally from future frames using geometric temporal
offsets or from other trajectories, without treating cross-trajectory
goals as necessarily unreachable.
The upper expectile emphasizes higher-return transitions in the data.
We interpret the learned score $R^{\mathcal D}$ as a proxy for
\emph{dataset-supported reachability}, assuming downstream controller
capabilities broadly compatible with the recorded behaviors.

\subsection{Task Utility Value Function}
\label{sec:utility}

Task utility measures progress toward the language-specified objective.
We adapt RoboMeter~\cite{liang2026robometer} with LoRA~\cite{hu2021lora} for most tasks in this paper,
supplemented by progress estimates from VLM outputs and SAM object
masks~\cite{ravi2025sam} for tasks with explicit spatial requirements,
such as Language Table~\cite{lynch2023interactive}.
For successful demonstrations, we apply a short moving average,
normalize scores to $[0,1]$ using start and completion anchors,
and take a running maximum.
The resulting annotation score $U(s_t)$ records the best progress
so far, with $U(s_0)=0$ and $U(s_{L-1})=1$.

\subsection{Subgoal Selection by Dynamic Programming}
\label{sec:method:dp}

For a demonstration $(s_0,\ldots,s_T)$ with $T=L-1$, we select
$K$ indices $0=t_0<t_1<\cdots<t_K\le T$, subject to
$t_k-t_{k-1}\ge\delta$.
Since consecutive utility gains telescope to $U(s_{t_K})$,
the objective is:
\begin{equation}
\min_{t_{1:K}}\sum_{k=1}^{K}
\bigl[w_r c_R(t_{k-1},t_k)+w_s\Omega_k(t_{k-1},t_k)\bigr]-w_g U(s_{t_K}),
\label{eq:dp}
\end{equation}
where
$c_R(i,j)=-\log\operatorname{clip}(R^{\mathcal D}(s_i,s_j),\epsilon,1)$
penalizes difficult transitions.
The spacing prior is $\Omega_k(i,j)=((j-i)/\mu_k(i)-1)^2$,
with $\mu_k(i)=(T-i)/(K-k+2)$ accounting for the remaining selected
subgoals and an appended terminal frame.
Utility favors a high-progress endpoint, reachability guides
intermediate placement, and the additional spacing regularizes temporal gaps.

Let $J_m(i)$ be the minimum remaining cost after frame $i$ with
$m$ subgoals left, and define
$e_m(i,j)=w_r c_R(i,j)+w_s\Omega_{K-m+1}(i,j)$.
The exact backward recurrence is
\begin{equation}
\begin{aligned}
J_m(i)&=\min_j\bigl[e_m(i,j)+J_{m-1}(j)\bigr],\\
J_0(i)&=-w_g U(s_i),
\end{aligned}
\label{eq:dp-recursion}
\end{equation}
where $i+\delta\le j\le T-(m-1)\delta$ and infeasible states
have infinite cost.
Filling the table takes $O(KL^2)$ time after score computation;
backtracking the stored minimizers from $J_K(0)$ recovers the
selected frames.
Appending $s_T$ gives the planner targets
$g_{1:N}=(s_{t_1},\ldots,s_{t_K},s_T)$, where $N=K+1$. Figure \ref{fig:method} illustrates this process.

\subsection{Subgoal Generation with a Pretrained Video Model}
\label{sec:method}

We finetune a pretrained video generation model, Wan2.2-TI2V-5B~\cite{wan2025wan}, as the generative planner
$\mathcal{P}_{\theta}$ to predict
subgoal sequence from the current observation and instruction. 

\paragraph{Sparse latent construction}
Wan2.2 operates on latents
produced by a causal 3D VAE with temporal compression factor $4$.
Directly encoding sparse subgoal images would mix neighboring
subgoals within the same temporal latent. To preserve a one-to-one
correspondence between subgoals and latent slots, we construct the input
sequence as:
\begin{equation}
\bigl[
o_t,
\underbrace{g_1,\ldots,g_1}_{4},
\ldots,
\underbrace{g_K,\ldots,g_K}_{4}
\bigr]
\xrightarrow{\mathrm{VAE}}
\bigl[
z_{o_t},z_{g_1},\ldots,z_{g_K}
\bigr].
\label{eq}
\end{equation}
Thus, each subgoal is compressed to a distinct latent. 

\paragraph{Closed-loop planner fine-tuning}
We train the DiT with LoRA~\cite{hu2021lora} using the model's
flow-matching objective, while freezing the VAE and text encoder.
Conditioning observations are sampled throughout each trajectory and
paired with subsequent annotated subgoals, exposing the planner to
intermediate task states. This supports periodic replanning from the
latest observation during execution.
For domains with limited language diversity, we apply instruction
dropout with probability $0.5$ to encourage visual grounding.

\paragraph{Prefix-conditioned generation for test-time search.}
To support test-time search and partial plan regeneration, we train the planner with \emph{random clean-prefix conditioning}. With probability $0.3$, we use standard single-observation conditioning; otherwise, we sample
$P\sim\mathrm{Uniform}_{1,\ldots,K-4}$ and condition on the clean latent prefix
$[z_{o_t},z_{g_1},\ldots,z_{g_P}]$, while applying the denoising loss only to the remaining subgoals
$[z_{g_{P+1}},\ldots,z_{g_K}]$.
Prefix positions are assigned diffusion timestep $0$ using a clean-frame mask.
With the causal VAE, this matches the prefix representation used at inference time, enabling the planner to retain selected subgoals and regenerate only the remaining plan during test-time search.

\subsection{Goal-Conditioned Low-Level Policy}
\label{sec:method}

The low-level controller $\pi_\phi$ is a small (0.1B parameters) goal-conditioned flow-matching Transformer that maps the current visual observations, proprioceptive state, language instruction $\ell$, and the generated sparse plan $\hat{g}_{1:K}$ to a chunk of low-level robot actions. The policy conditions on several future subgoals rather than only the next image. Visual observations and subgoals are represented in a shared frozen RADIO feature space~\cite{ranzinger2024radio}, while the subgoal sequence is compressed into a small set of tokens before being fused with proprioceptive and language features. The resulting conditioning tokens form a prefix to a flow-matching Transformer that generates the action chunk.
\section{Experiments}
\label{sec:exp}

Our experiments are designed to systematically evaluate the key advantages of \name{} along the following dimensions:
(1)~\emph{Long-horizon task planning capability} (\S\ref{sec:exp:long-horizon});
(2)~\emph{Hierarchical task execution effectiveness} (\S\ref{sec:exp:hierarchical}); (3)~\emph{Scaling with sub-optimal dataset} (\S\ref{sec:exp:suboptimal}); (4)~\emph{Test-time search and language steering} (\S\ref{sec:exp:test-time}); (5)~\emph{Subgoal Reachability} (\S\ref{sec:exp:staged});(6)~\emph{Real-world applicability} (\S\ref{sec:exp:real}). The experimental settings are introduced in each part, respectively.

\begin{figure}
    \centering
    \includegraphics[width=\linewidth]{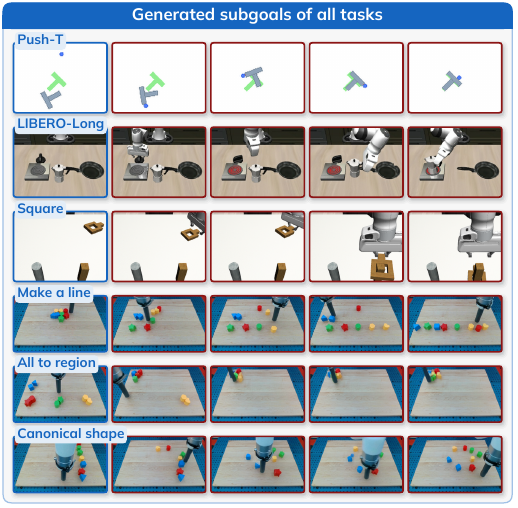}
    \caption{Illustration of the subgoals generated from our methods on all simulation tasks in this paper.}
    \label{fig:illustration}
\end{figure}

\subsection{Long-horizon Task Planning Results}
\label{sec:exp:long-horizon}

We choose three tasks from the long-horizon real-world version of the Language Table~\cite{lynch2023interactive} dataset: 1) \textit{make a line}: the robot needs to place the randomly initialized blocks on the table into a line according to the language instruction; 2) \textit{all to region}: the robot needs to cluster blocks into different regions according to the language instruction; 3) \textit{canonical shape}: the robot needs to arrange the blocks into specified standard shapes according to the language instruction. We use 1,424, 1,303, and 1,378 videos, respectively, for the training of the three tasks, and 100 videos for testing for all three tasks. We resize each video to $320\times 240$ resolution for training, and set 15 subgoals for each episode.

A plan counts as successful only if it jointly satisfies goal attainment (the final frame forms the instructed target shape), object integrity (no block vanishes, splits, or is hallucinated for a sustained stretch), and transition validity (no tracked block displaces implausibly). For the baseline, we choose VLP~\cite{vlp}, which is a method that uses language as a planning representation and uses a video generation model to imagine the next state. We choose it to demonstrate the advantage of visual planning using our method.

Table~\ref{tab:language_table} shows the results. We can see that our method consistently outperforms VLP on all three tasks. This demonstrates the advantage of visual planning over language planning. A major failure mode of VLP is that language instructions often cannot precisely specify the scene changes required for task completion. For example, given the instruction “\textit{put the green star towards the green circle}”, the robot is not told exactly where the green star should be placed relative to the green circle, nor how it should react if the star encounters other blocks along the way. We illustrate our planning results in Figure \ref{fig:illustration}.

\begin{table}[t]
    \centering
    \caption{Visual plan generation quality comparisons on Language Table (LT) and LIBERO-Long (LL) tasks.}
    \vspace{-0.05in}
    \label{tab:visual}
    \resizebox{\linewidth}{!}{
    \begin{tabular}{l|cc|cc|cc|cc}
    \toprule
    \textbf{Method} 
    & \multicolumn{2}{c|}{PSNR $\uparrow$} 
    & \multicolumn{2}{c|}{SSIM $\uparrow$} 
    & \multicolumn{2}{c|}{LPIPS $\downarrow$} 
    & \multicolumn{2}{c}{FVD $\downarrow$} \\
    \cmidrule(lr){2-3}
    \cmidrule(lr){4-5}
    \cmidrule(lr){6-7}
    \cmidrule(lr){8-9}
    & LT & LL & LT & LL & LT & LL & LT & LL \\
    \midrule
    Dense Visual Planning 
    & 27.54 & 18.52 & 0.859 & 0.834 & 0.115 & 0.128 & 233 & 370 \\
    Uniform Subgoal Planning 
    & 27.51 & 18.06 & 0.855 & 0.826 & 0.118 & 0.143 & 258 & 380 \\
    Ours 
    & \textbf{28.95} & \textbf{18.86} & \textbf{0.870} & \textbf{0.838} & \textbf{0.103} & \textbf{0.127} & \textbf{154} & \textbf{363} \\
    \bottomrule
    \end{tabular}}
\end{table}

We also compare our method with different planning strategies on the \textbf{make a line} task with: 1) \textbf{dense video planning}: we fine-tune WAN 2.2~\cite{wan2025wan} to generate the future dense videos, serving as a standard video generation planning baseline; 2) \textbf{Uniform subgoal planning}: we uniformly sample 15 subgoals on each training video and finetune WAN 2.2 to generate the subgoal sequence. This is to demonstrate the advantage of our subgoal discovery method. Results are shown in the rightmost column of Table \ref{tab:main_results}. We can see that our method is better than other planning methods. Dense planning performs poorly because long-horizon tasks require autoregressive generation of long video sequences, leading to severe hallucinations such as object disappearance and deformation. This is a common issue in world-model-based approaches. Uniform subgoal planning suffers from suboptimal behaviors in the dataset. Uniform sampling cannot skip unnecessary suboptimal steps, making task progress non-monotonic. This confuses the video generation model and often leads to implausible jumps in generated subgoals.

We also evaluate the pixel-level and feature-level quality of the generated visual plans. Results are in Table~\ref{tab:visual}. We can see that our method leads on all metrics.

\begin{table}[t]
    \centering
    \caption{Long-horizon task planning success rates on three Language Table tasks. We reproduced the \textit{canonical shape} result by ourselves because the original VLP didn't do it.}
    \resizebox{0.75\linewidth}{!}{
    \begin{tabular}{c|cc}
    \toprule
    & VLP~\cite{vlp} & \name{} (Ours)\\
    \midrule
    make a line & 56\%& \textbf{86\%}\\
    all to region & 78\%& \textbf{100\%} \\ 
    canonical shape & 15\%& \textbf{52\%}  \\
    \bottomrule
    \end{tabular}}
    \label{tab:language_table}
\end{table}

\begin{table}[t]
    \centering
    \caption{Comparison of different planning methods on PushT, LIBERO-Long, and Language Table \textit{make a line} task. Numbers for our methods are the average success rates of 50 evaluation episodes. The VLA and WAM results on PushT were run by us. LIBERO-Long results are from their original papers. Bold numbers indicate the best results or our results.}
    \label{tab:main_results}
    \resizebox{\linewidth}{!}{
    \begin{tabular}{lccc}
        \toprule
        \textbf{Method} & \textbf{PushT} & \textbf{LIBERO-Long} &
        \textbf{Language Table}\\
        \midrule
        E2E Policy               & 58.0\% & 79.2\% & --\\
        Language Planning        & 62.0\% & 75.6\% & 56.0\%\\
        Dense Visual Planning    & 70.0\% & 84.0\% & 71.0\%\\
        Uniform Subgoal Planning  & 72.0\% & 82.4\% & 69.0\%\\
        \midrule
        VLA-OS~\cite{vlaos} (0.9B, non-pretrained) & 72.0\% & 76.0\% & --\\
        $\pi_{0}$~\cite{black2024pi0} (3.3B, pretrained) & 76.0\% & 85.2\% & --\\            
        SEER~\cite{seer} (0.3B, pretrained) & 68.0\% & 87.7\% & --\\            $\pi_{0.5}$~\cite{intelligence2025pi05} (3.3B, pretrained) & 86.0\% & 92.4\% & --\\
        Lingbot-VA~\cite{lingbotva} (5.3B, pretrained) & \textbf{89.0\%} & \textbf{98.5\%} & --\\
        \midrule
        \textbf{Ours} (0.1B policy, non-pretrained)           & \textbf{86.0\%} & \textbf{91.2\%} & \textbf{86.0\%}\\
        \bottomrule
    \end{tabular}
    }
\end{table}

\subsection{Hierarchical Task Execution Results}
\label{sec:exp:hierarchical}

We use PushT~\cite{florence2022implicit} and LIBERO-Long~\cite{liu2023libero} to demonstrate the effectiveness of the complete hierarchical system, pairing our planner with a goal-conditioned low-level policy, as measured by task success rates. We use the official 206 demonstrations to train the PushT task, and 379 demonstrations to train LIBERO-Long. We use the same low-level policy architecture across our method and all baselines.

Results are shown in Table \ref{tab:main_results}. We can see that, with the policy, budget, and protocol
held fixed, the planning representation alone separates the methods: language planning
\emph{underperforms} even the end-to-end baseline; dense video planning helps, uniform
subgoals help more, but they are all worse than \name{}'s value-selected subgoals.

We also compare our method with several SOTA large models, including two VLAs ($\pi_0$~\cite{black2024pi0} and $\pi_{0.5}$~\cite{intelligence2025pi05}), a hierarchical VLA model (VLA-OS~\cite{vlaos}), and two world-action-models (SEER~\cite{seer} and Lingbot-VA~\cite{lingbotva}). Results are in Table \ref{tab:main_results}. We can see that our method surpasses both the non-pretrained image subgoal planner (VLA-OS), the small pretrained video latent planner (SEER), and pretrained end-to-end VLA models ($\pi_0$). Although our method is still no better than the SOTA large pretrained VLA ($\pi_{0.5}$) or WAM model (Lingbot-VA), this result can already demonstrate that even a small 0.1B non-pretrained policy can be comparable to large pretrained VLA or WAM models as long as helpful subgoal images. 

\subsection{Scaling
with Sub-Optimal Dataset}
\label{sec:exp:suboptimal}

We use Robomimic-MH Square~\cite{robomimic} to demonstrate our method can scale with suboptimal demonstrations. This task has 300
demonstrations graded by operator skill into 100 proficient
and 200 suboptimal. We first train a goal-conditioned low-level policy with 100 optimal demonstrations, where the goal image is selected randomly from the same episode after the current image. Then we build eight nested subsets of the suboptimal pool ($N \in \{5,10,20,40,80,120,160,200\}$, where each subset contains the previous, then train our subgoal generator per subset and evaluate the hierarchical success rate. We also train a set of dense video planners and uniform subgoal planners under the same setting.

\begin{figure}
    \centering
    \includegraphics[width=\linewidth]{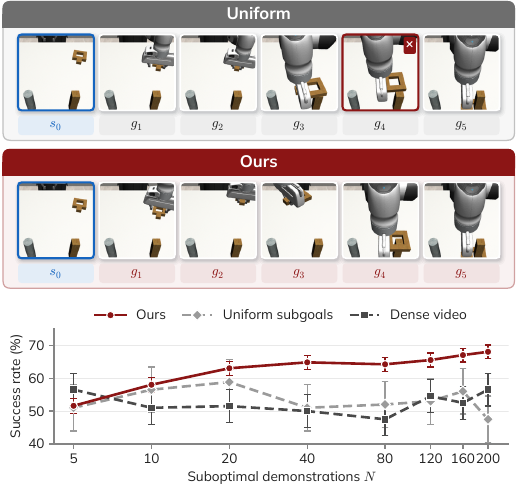}
    \caption{Top: subgoals found by uniform sampling and ours. Bottom: success rates on the Square task with more suboptimal data for different planners.}
    \label{fig:suboptimal}
\end{figure}

Results are shown in Figure \ref{fig:suboptimal}. We can see that \name\ can benefit from more and more suboptimal datasets, which is suitable for large-scale pretraining. In contrast, the dense video planner's performance increases slowly as more training data is added, while the uniform subgoal planner's performance fluctuates due to its incorrect subgoals.


\begin{figure*}[t]
    \centering
    \includegraphics[width=\linewidth]{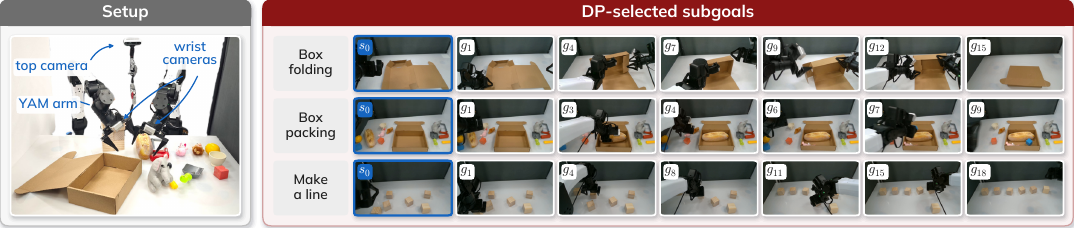}
    \caption{The real-world experiment setup, three real-world tasks, and subgoal sequences.}
    \label{fig:real_world}
\end{figure*}

\subsection{Test-Time Search and Language Steering}
\label{sec:exp:test-time}

\begin{figure}
    \centering
    \includegraphics[width=\linewidth]{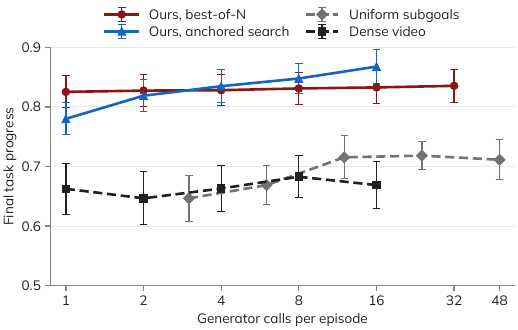}
    \caption{Test time search results on the Language Table \textit{make a line} task. Each point is evaluated over 100 testing episodes.}
    \label{fig:tts}
\end{figure}

\begin{figure}
    \centering
    \includegraphics[width=\linewidth]{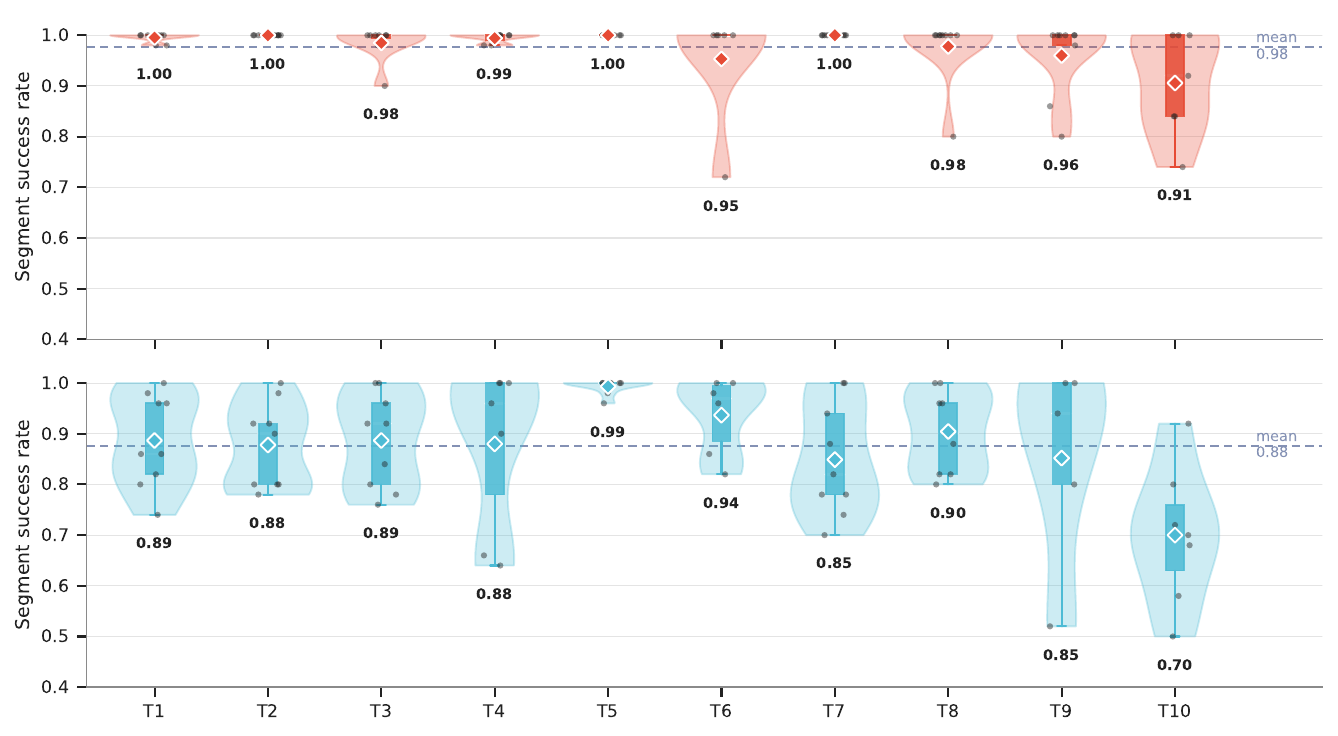}
    \caption{Per-subgoal reachability on 10 LIBERO-Long tasks. Top: ours. Bottom: uniform subgoal planning.}
    \label{fig:per-subgoal}
\end{figure}

\begin{figure*}[t]
    \centering
    \includegraphics[width=1.0\linewidth]{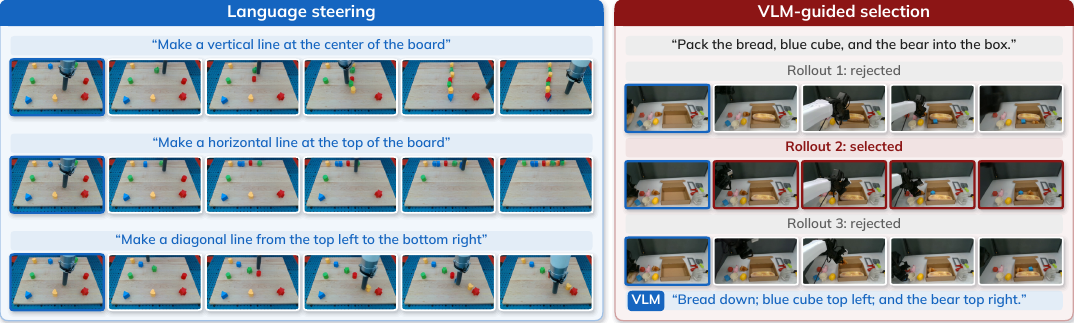}
    \caption{Language steering experiments. Left: given different language instructions, our planner can generate different corresponding planning results. Right: given only the black instruction, our planner can generate different rollouts, while a VLM instructed by the green text can give a task progress value function that selects the correct planning rollout.}
    \label{fig:steering}
\end{figure*}

We use the Language Table \textit{make a line} task as well as the real-world \textit{box packing} task to demonstrate the test-time search ability on our generated subgoal sequences. First, we show on \textit{make a line} task that the planning task progress can be increased with more test-time computing added, as shown in Figure \ref{fig:tts}. Here, we use the same task progress value function during training for test-time search, which is a scripted, detection-based score. We compare two search methods:
(i) whole-sequence best-of-$N$; (ii) anchored search: at each round, it samples candidates for the remaining subgoals, commits the best
$c$, pins them as clean latents, and regenerates the tail. We also run dense video planner with
the identical value function and matched compute on best-of-$N$ search. We can see that our method can effectively benefit from anchored search. This is viable thanks to the multi-anchor training. The dense video planner also benefits from test-time search a little, but the final performance is still worse than the initial results of our planner.

Next, we show that given different language instructions, our planner can be steered accordingly. In Figure \ref{fig:steering}, given different language instructions, we first use VLMs to write corresponding task progress value functions, and then use them to steer the generated image sequences. We can see that our method can generate different rollouts given different language instructions from the same initial input.

\subsection{Reachability Value Function Evaluation}
\label{sec:exp:staged}

We use LIBERO-Long tasks to illustrate the reachability of each subgoal discovered by our method. For
every consecutive ground-truth subgoal pair $(s_i, s_{i+1})$, the simulator is reset and replayed
to $s_i$, control is handed to \name{}, and the segment counts as a success if the
task-relevant objects reach their $s_{i+1}$ configuration within $5$ cm. Results are shown in Figure~\ref{fig:per-subgoal}. We can see that our method can achieve per-segment success rates near the ceiling.

We also conduct ablation studies to show the effectiveness of different dynamic programming terms on LIBERO-Long. We test using spacing only (S), reachability and spacing (RS), utility and spacing (US), reachability and utility (RU), and ours, to discover subgoals, and train the planner and low-level policy in the same way. Results are shown in Table \ref{tab:ablation_terms}. We can see that with only S or without S, the result is similar to uniform planning; without R or U, the results are similar to the end-to-end policy, showing that the subgoals are not correct. This shows that our three terms are all necessary.

\begin{table}[t]
    \centering
    \caption{The ablation study of subgoal discovery terms.}
    \resizebox{0.8\linewidth}{!}{
    \begin{tabular}{c|ccccc}
    \toprule
      DP Terms   &  S & RS & US & RU & Ours \\
    \midrule
       SR  & 81.8\% & 74.2\% & 78.0\% & 83.8\% & \textbf{91.2\%} \\ 
    \bottomrule
    \end{tabular}}
    \label{tab:ablation_terms}
\end{table}

\begin{figure}
    \centering
    \includegraphics[width=\linewidth]{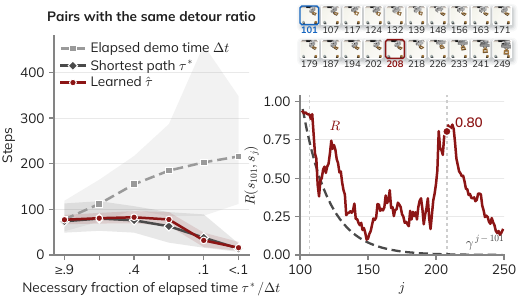}
    \caption{Left: learned reachability follows the shortest path, not elapsed time. Right: our reachability skips a failed grasp.}
    \label{fig:r}
\end{figure}

To further show that our reachability function is not a simple elapsed timestep approximation, we show in the left part of Figure \ref{fig:r} of frame pairs from the 200 sub-optimal Square demos, binned by the fraction of elapsed demo time that was necessary, $\tau^*/\Delta t$, where $\tau^*$ is the shortest-path hitting time over all dataset frames. As demonstrations detour more (right), elapsed time $\Delta t$ keeps growing while $\tau^*$ shrinks; the time-to-reach implied by the learned value, $\hat{\tau}$, tracks $\tau^*$. A reachability that merely decayed with time would follow the dashed line. Medians with inter-quartile bands. We also show on the right with a sub-optimal Square demo where the nut is grasped at $t=101$, dropped at $107$, and re-grasped at $208$. Anchored at $s_{101}$, the learned reachability $R(s_i, s_j)$ (red) falls while first, then rebounds to 0.80 at the regrasp, where the state is back within 1.4cm of $s_i$. A time-decay prior $\gamma^{j-i}$ (dashed) gives 0.004.

\subsection{Real-World Evaluation}
\label{sec:exp:real}

We further evaluate \name{} on three long-horizon real-world tasks using a dual-arm YAM robot, with one RealSense D435 camera as the top-view camera and two RealSense D405 cameras as wrist cameras, as shown in Figure~\ref{fig:real_world}: 1) \textbf{Cardboard Box Folding}: a contact-rich task where the robot needs to fold the unfolded cardboard box into its usable configuration for the randomly placed cardboard box. We give 100 human teleoperation demonstrations for this task; 2) \textbf{Box Packing}: a combination generalization task where the robot needs to follow a language instruction to pack a specified set of objects from the tabletop into the box. There are 2 big objects, 3 medium objects, and 6 small objects, and a total of 205 feasible packing combinations. We choose 110 combinations for training and give 5 demonstrations for each of them, and leave the other 95 combinations for zero-shot evaluation; 3) \textbf{Make a Line}: the robot needs to move the six cubes into a line. We give 90 demonstrations across 3 different kinds of target line shapes. 

Results are shown in Table~\ref{tab:real_world}. We can see that our method outperforms other planning methods.

\begin{table}[t]
    \centering
    \caption{Real-world experiment results. We evaluate each task with 20 episodes.}
    \resizebox{\linewidth}{!}{
    \begin{tabular}{lccc}
    \toprule
    Method & Box Folding & Box Packing (Unseen) & Make a Line\\
    \midrule
    End-to-End Policy& 5\% & 0\% & 10\%\\
    Dense Video Planning & 0\% & 5\% & 0\%\\
    Uniform Subgoal Planning & 10\% & 20\% & 25\%\\
    Ours & \textbf{40\%} & \textbf{65\%} & \textbf{55\%}\\
    \bottomrule
    \end{tabular}}
    \label{tab:real_world}
\end{table}

\section{Conclusion}
We presented \name, a sparse visual planning framework for long-horizon robot manipulation that represents plans as task-relevant subgoal images. By balancing dataset-supported \emph{reachability} and \emph{task utility}, \name{} automatically discovers subgoals and fine-tunes a pretrained video model into a sparse visual planner. Across simulated and real-world tasks, \name{} outperforms language, dense-video, and naive subgoal-planning baselines, while having superior hierarchical system performance, supporting test-time search and language steering, suboptimal-data scaling, and efficient closed-loop replanning.


{\small
\bibliographystyle{IEEEtran}
\bibliography{references}

@article{akgun2012keyframe,
  title   = {Keyframe-based Learning from Demonstration},
  author  = {Akgun, Baris and Cakmak, Maya and Jiang, Karl and Thomaz, Andrea L.},
  journal = {IJSR},
  year    = {2012},
}

@article{gregor2016variational,
  title   = {Variational Intrinsic Control},
  author  = {Gregor, Karol and Rezende, Danilo Jimenez and Wierstra, Daan},
  journal = {arXiv preprint arXiv:1611.07507},
  year    = {2016}
}

@inproceedings{eysenbach2019diversity,
  title     = {Diversity is All You Need: Learning Skills without a Reward Function},
  author    = {Eysenbach, Benjamin and Gupta, Abhishek and Ibarz, Julian and Levine, Sergey},
  booktitle = {ICLR},
  year      = {2019}
}

@inproceedings{pertsch2020keyin,
  title     = {Keyframing the Future: Keyframe Discovery for Visual Prediction and Planning},
  author    = {Pertsch, Karl and Rybkin, Oleh and Yang, Jingyun and Zhou, Shenghao and Derpanis, Konstantinos G. and Daniilidis, Kostas and Lim, Joseph J. and Jaegle, Andrew},
  booktitle = {L4DC},
  year      = {2020},
  organization = {PMLR}
}

@article{wang2024vlm,
  title   = {VLM See, Robot Do: Human Demo Video to Robot Action Plan via Vision Language Model},
  author  = {Wang, Beichen and Zhang, Juexiao and Dong, Shuwen and Fang, Irving and Feng, Chen},
  journal = {arXiv preprint arXiv:2410.08792},
  year    = {2024}
}

@article{hafner2023dreamerv3,
  title   = {Mastering Diverse Domains through World Models},
  author  = {Hafner, Danijar and Pasukonis, Jurgis and Ba, Jimmy and Lillicrap, Timothy},
  journal = {arXiv:2301.04104},
  year    = {2023}
}

@inproceedings{hansen2024tdmpc2,
  title     = {TD-MPC2: Scalable, Robust World Models for Continuous Control},
  author    = {Hansen, Nicklas and Su, Hao and Wang, Xiaolong},
  booktitle = {ICLR},
  year      = {2024}
}

@inproceedings{du2023unipi,
  title     = {Learning Universal Policies via Text-Guided Video Generation},
  author    = {Du, Yilun and Yang, Sherry and Dai, Bo and Dai, Hanjun and Nachum, Ofir and Tenenbaum, Joshua B. and Abbeel, Pieter and Schuurmans, Dale},
  booktitle = {NeurIPS},
  year      = {2023}
}

@inproceedings{yang2024unisim,
  title     = {Learning Interactive Real-World Simulators},
  author    = {Yang, Mengjiao and Du, Yilun and Ghasemipour, Kamyar and Tompson, Jonathan and Kaelbling, Leslie and Schuurmans, Dale and Abbeel, Pieter},
  booktitle = {International Conference on Learning Representations (ICLR)},
  year      = {2024}
}

@inproceedings{li2025unifiedvideoaction,
  title     = {Unified Video Action Model},
  author    = {Li, Shuang and Gao, Yihuai and Sadigh, Dorsa and Song, Shuran},
  booktitle = {Robotics: Science and Systems (RSS)},
  year      = {2025}
}

@article{ye2026dreamzero,
  title   = {World Action Models are Zero-shot Policies},
  author  = {Ye, Seonghyeon and Ge, Yunhao and Zheng, Kaiyuan and Gao, Shenyuan and Yu, Sihyun and Kurian, George and Indupuru, Suneel and Tan, You Liang and Zhu, Chuning and Xiang, Jiannan and others},
  journal = {arXiv preprint arXiv:2602.15922},
  year    = {2026}
}

@article{saycan,
  title={Do as i can, not as i say: Grounding language in robotic affordances},
  author={Ahn, Michael and Brohan, Anthony and Brown, Noah and Chebotar, Yevgen and Cortes, Omar and David, Byron and Finn, Chelsea and Fu, Chuyuan and Gopalakrishnan, Keerthana and Hausman, Karol and others},
  journal={arXiv preprint arXiv:2204.01691},
  year={2022}
}

@article{iql,
  title={Offline reinforcement learning with implicit q-learning},
  author={Kostrikov, Ilya and Nair, Ashvin and Levine, Sergey},
  journal={arXiv preprint arXiv:2110.06169},
  year={2021}
}

@article{park2023hiql,
  title={Hiql: Offline goal-conditioned rl with latent states as actions},
  author={Park, Seohong and Ghosh, Dibya and Eysenbach, Benjamin and Levine, Sergey},
  journal={NeurIPS},
  year={2023}
}

@article{liang2026robometer,
  title={Robometer: Scaling general-purpose robotic reward models via trajectory comparisons},
  author={Liang, Anthony and Korkmaz, Yigit and Zhang, Jiahui and Hwang, Minyoung and Anwar, Abrar and Kaushik, Sidhant and Shah, Aditya and Huang, Alex S and Zettlemoyer, Luke and Fox, Dieter and others},
  journal={arXiv preprint arXiv:2603.02115},
  year={2026}
}

@article{wan2025wan,
  title={Wan: Open and advanced large-scale video generative models},
  author={Wan, Team and Wang, Ang and Ai, Baole and Wen, Bin and Mao, Chaojie and Xie, Chen-Wei and Chen, Di and Yu, Feiwu and Zhao, Haiming and Yang, Jianxiao and others},
  journal={arXiv preprint arXiv:2503.20314},
  year={2025}
}

@article{liu2023libero,
  title={Libero: Benchmarking knowledge transfer for lifelong robot learning},
  author={Liu, Bo and Zhu, Yifeng and Gao, Chongkai and Feng, Yihao and Liu, Qiang and Zhu, Yuke and Stone, Peter},
  journal={NeurIPS},
  year={2023}
}

@inproceedings{florence2022implicit,
  title={Implicit behavioral cloning},
  author={Florence, Pete and Lynch, Corey and Zeng, Andy and Ramirez, Oscar A and Wahid, Ayzaan and Downs, Laura and Wong, Adrian and Lee, Johnny and Mordatch, Igor and Tompson, Jonathan},
  booktitle={CoRL},
  year={2022},
}

@article{lynch2023interactive,
  title={Interactive language: Talking to robots in real time},
  author={Lynch, Corey and Wahid, Ayzaan and Tompson, Jonathan and Ding, Tianli and Betker, James and Baruch, Robert and Armstrong, Travis and Florence, Pete},
  journal={IEEE Robotics and Automation Letters},
  year={2023},
  publisher={IEEE}
}

@article{black2024pi0,
  title={pi0: A Vision-Language-Action Flow Model for General Robot Control},
  author={Intelligence, Physical},
  journal={arXiv},
  year={2024}
}

@article{intelligence2025pi05,
  title={pi05: a Vision-Language-Action Model with Open-World Generalization},
  author={Intelligence, Physical},
  journal={arXiv preprint arXiv:2504.16054},
  year={2025}
}

@inproceedings{seer,
  title={Predictive inverse dynamics models are scalable learners for robotic manipulation},
  author={Tian, Yang and Yang, Sizhe and Zeng, Jia and Wang, Ping and Lin, Dahua and Dong, Hao and Pang, Jiangmiao},
  booktitle={ICLR},
  year={2025}
}

@article{robomimic,
  title={What matters in learning from offline human demonstrations for robot manipulation},
  author={Mandlekar, Ajay and Xu, Danfei and Wong, Josiah and Nasiriany, Soroush and Wang, Chen and Kulkarni, Rohun and Fei-Fei, Li and Savarese, Silvio and Zhu, Yuke and Mart{\'\i}n-Mart{\'\i}n, Roberto},
  journal={arXiv preprint arXiv:2108.03298},
  year={2021}
}

@misc{wgobench,
  author       = {{Macrodata Labs}},
  title        = {Segmenting Robot Video into Actionable Subtasks},
  year         = {2026},
  howpublished = {\url{https://macrodata.co/blog/annotating-robot-video-subtasks}},
}

@inproceedings{liu2025lbp,
  title     = {Efficient Robotic Policy Learning via Latent Space Backward Planning},
  author    = {Liu, Dongxiu and Niu, Haoyi and Wang, Zhihao and Zheng, Jinliang
               and Zheng, Yinan and Ou, Zhonghong and Hu, Jianming
               and Li, Jianxiong and Zhan, Xianyuan},
  booktitle = {International Conference on Machine Learning},
  year      = {2025}
}

@inproceedings{black2024susie,
  title     = {Zero-Shot Robotic Manipulation with Pretrained Image-Editing
               Diffusion Models},
  author    = {Black, Kevin and Nakamoto, Mitsuhiko and Atreya, Pranav
               and Walke, Homer and Finn, Chelsea and Kumar, Aviral
               and Levine, Sergey},
  booktitle = {ICLR},
  year      = {2024}
}

@article{yuan2026fastwam,
  title   = {Fast-WAM: Do World Action Models Need Test-time Future Imagination?},
  author  = {Yuan, Tianyuan and Dong, Zibin and Liu, Yicheng and Zhao, Hang},
  journal = {arXiv},
  year    = {2026}
}

@inproceedings{ranzinger2024radio,
  title={Am-radio: Agglomerative vision foundation model reduce all domains into one},
  author={Ranzinger, Mike and Heinrich, Greg and Kautz, Jan and Molchanov, Pavlo},
  booktitle={CVPR},
  year={2024},
  organization={IEEE}
}

@article{hu2021lora,
  title={Lora: Low-rank adaptation of large language models},
  author={Hu, Edward J and Shen, Yelong and Wallis, Phillip and Allen-Zhu, Zeyuan and Li, Yuanzhi and Wang, Shean and Wang, Lu and Chen, Weizhu},
  journal={arXiv preprint arXiv:2106.09685},
  year={2021}
}

@inproceedings{vlp,
  title={Video language planning},
  author={Du, Yilun and Yang, Sherry and Florence, Pete and Xia, Fei and Tenenbaum, Joshua B and Kaelbling, Leslie and others},
  booktitle={ICLR},
  year={2024}
}

@article{vlaos,
  title={VLA-OS: Structuring and dissecting planning representations and paradigms in vision-language-action models},
  author={Gao, Chongkai and Liu, Zixuan and Chi, Zhenghao and Huang, Junshan and Fei, Xin and Hou, Yiwen and Zhang, Yuxuan and Lin, Yudi and Fang, Zhirui and Shao, Lin},
  journal={NeurIPS},
  year={2026}
}

@article{lingbotva,
  title={Causal world modeling for robot control},
  author={Li, Lin and Zhang, Qihang and Luo, Yiming and Yang, Shuai and Wang, Ruilin and Han, Fei and Yu, Mingrui and Gao, Zelin and Xue, Nan and Zhu, Xing and others},
  journal={arXiv preprint arXiv:2601.21998},
  year={2026}
}

@inproceedings{ravi2025sam,
  title={Sam 2: Segment anything in images and videos},
  author={Ravi, Nikhila and Gabeur, Valentin and Hu, Yuan-Ting and Hu, Ronghang and Ryali, Chaitanya and Ma, Tengyu and Khedr, Haitham and R{\"a}dle, Roman and Rolland, Chloe and Gustafson, Laura and others},
  booktitle={ICLR},
  year={2025}
}

@inproceedings{zhou2026gem,
  title={Gem-4d: Geometry-enhanced video world models for robot manipulation},
  author={Zhou, Kaichen and Chen, Yuzhen and Zhan, Fangneng and Hua, Hang and Chen, Grace and Chang, Xinhai and Qu, Ao and Du, Yilun and Liu, Zhuang and Liang, Paul Pu and others},
  booktitle={European Conference on Computer Vision},
  pages={130--147},
  year={2026},
  organization={Springer}
}

@inproceedings{gao2025flip,
  title={Flip: Flow-centric generative planning as general-purpose manipulation world model},
  author={Gao, Chongkai and Zhang, Haozhuo and Xu, Zhixuan and Zhehao, Cai and Shao, Lin},
  booktitle={International Conference on Learning Representations},
  volume={2025},
  pages={21927--21948},
  year={2025}
}
}


\end{document}